# Evaluating the point cloud of individual trees generated from images based on Neural Radiance fields (NeRF) method

Hongyu Huang[1,2,3,*], Guoji Tian[1,2,3], Chongcheng Chen[1,2,3],

1. National Engineering Research Center of Geospatial Information Technology, Fuzhou University, Fuzhou 350108, China;
2. Key Lab of Spatial Data Mining and Information Sharing of Ministry of Education, Fuzhou University, Fuzhou 350108, China
3. The Academy of Digital China (Fujian), Fuzhou 350108, China
    * **Corresponding author:** E-mail: hhy1@fzu.edu.cn; Tel.:+8613328269460

**Abstract**：Three-dimensional (3D) reconstruction of trees has always been a key task in precision forestry management and research. Due to the complex branch morphological structure of trees themselves and the occlusions from tree stems, branches and foliage, it is difficult to recreate a complete three-dimensional tree model from a two-dimensional image by conventional photogrammetric methods. In this study, based on tree images collected by various cameras in different ways, the Neural Radiance Fields (NeRF) method was used for individual tree reconstruction and the exported point cloud models are compared with point cloud derived from photogrammetric reconstruction and laser scanning methods. The results show that the NeRF method performs well in individual tree 3D reconstruction, as it has higher successful reconstruction rate, better reconstruction in the canopy area, it requires less amount of images as input. Compared with photogrammetric reconstruction method, NeRF has significant advantages in reconstruction efficiency and is adaptable to complex scenes, but the generated point cloud tends to be noisy and low resolution. The accuracy of tree structural parameters (tree height and diameter at breast height) extracted from the photogrammetric point cloud is still higher than those of derived from the NeRF point cloud. The results of this study illustrate the great potential of NeRF method for individual tree reconstruction, and it provides new ideas and research directions for 3D reconstruction and visualization of complex forest scenes.

**Keywords**：3D reconstruction; Neural Radiance Field (NeRF); 3D Tree Modelling; Photogrammetry; Deep Learning; Individual Tree; Terrestrial Laser Scanning;

## 1. Introduction

Trees are an essential part of the Earth's ecosystem, providing numerous critical ecological services and influencing many environmental aspects such as soil conservation, climate regulation, and wildlife habitat [1]. A comprehensive understanding of the distribution and number of trees, as well as information on tree morphology and structure, is important for forestry and natural resource management, as well as environmental monitoring, characterization and protection. The rapid development of 3D digitization, reconstruction technology and artificial intelligence (AI) provides a new direction for tree monitoring and protection. 3D technology is applied to forestry by collecting data and then performing automatic modelling, and key information such as tree height, diameter at breast height (DBH), crown diameter and volume of trees can be obtained for forest inventory and management purposes [2].

To acquire 3D data of trees, a variety of means can be used, which can usually be categorized into terrestrial platforms and aerial platforms. Ground-based platforms include: terrestrial laser scanning (TLS), mobile laser scanning (MLS) and ground-based photogrammetry [3]. TLS has the highest geometric data quality of all sensors and platforms, while MLS collects data efficiently but with slightly less accuracy than TLS. Ground-based photogrammetry, on the other hand, processes photos collected from a variety of cameras, which is simple and easy to operate. Airborne platforms mainly include UAVs, helicopters, etc., which are capable of acquiring high spatial resolution data quickly, with an accuracy even comparable to that of ground-based acquisition systems.

Terrestrial laser scanners are often used for forest surveys and 3D trees reconstruction, as they are able to acquire accurate point cloud models of trees in a non-contact, active manner. However, TLS is relatively expensive and difficult to carry around, which prevents it from being used easily in dense forests. In the last decade, photogrammetric method that is simple and affordable to acquire 3D data had become widely used for forest inventory and vegetation modeling works [4]. Huang et al. used a consumer-grade hand-held camera to acquire images of a desert plant, and were able to reconstruct a point-cloud model with accuracy comparable to that of derived from TLS [5]; Kükenbrink et al. [6] evaluated the performance of using an action camera (GoPro) combined with Structure from Motion (SfM) [7] technique to acquire 3D point clouds of trees, and their point cloud model was of similar quality to that of LiDAR devices in terms of the DBH extraction results, but still fell short of the performance in other aspects. For this reason, many scholars have opened up new ideas to acquire forest point cloud data using a portable camera and a backpack laser scanner, combining these two types of data to extract tree height and DBH [8]; Balestra et al. fused the data from RGB camera, UAV camera and mobile laser scanner to reconstruct three giant chestnut tree models, providing researchers with accurate information about the shape and overall condition of the trees [9]. Although these photogrammetry-based tree reconstruction methods are simple, automated and effective, the reconstruction accuracy is greatly affected by the fact that the trees themselves have complex self-similar branch morphological structures and the branches are shaded by leaves. Current photogrammetry method can deal with trees with prominent stem and branches features that are not covered or concealed by the foliage, but for trees with dense leaves in their canopies, the reconstruction result is often not satisfactory: usually only the lower trunk or stem can be recovered, while the upper canopy is partly or totally missing in the 3D model.

After the concept of deep learning [10] was first introduced in 2006, it has been developed rapidly in various fields, including computer graphics and 3D reconstruction. Neural Radiance Fields (NeRF) [11] is one example of recent major development. NeRF create photorealistic 3D scenes from series photos. Given the input of a set of calibrated images, its goal is to output a volumetric 3D scene that renders novel views. NeRF's main task is to synthesize new view based on the known view images, but it also can be used to generate mesh using marching cubes; and further in nerfstudio [21, 26] the reconstructed mesh and point cloud can be exported. This technique not only enables the rendering of photo-quality views, but also is more adaptable to the rendering of texture-less and transparent objects compared to conventional photogrammetric reconstruction methods. Since NeRF was proposed, it has been extensively studied and used in various fields, including autonomous driving [12], medicine [13], digital human body [14], 3D cities [15], and cultural heritage reconstruction [16]. However, there has been no published research on NeRF for 3D reconstruction of trees so far, and the advantages and shortcomings of NeRF method have not been adequately studied, comparing with the relatively mature photogrammetric reconstruction methods. Therefore, in this paper, we applied NeRF technology to achieve 3D reconstruction of trees and used the derived point cloud for tree structural parameters extraction and 3D modeling. Our goal is to answer these questions: how good is the derived point cloud (both visually and quantitatively) compared with traditional photogrammetric reconstruction methods, and what are the strength and weakness of this method? We wish to use this study to collect information and gain new ideas and directions for the 3D reconstruction of trees.

## 2. Materials and Method

### 2.1. Study Area

We chose two trees located within the Qishan Campus of Fuzhou University for this study. Tree_1, a recently transplanted Autumn Maple tree, is located in a fairly open area; it is easy to take photos around the tree as the view was not being blocked or interfered with by other vegetative material. The tree has distinctive trunk characteristics, sparse foliage, and a relatively simple crown and branch structure. Tree_2 is an imposing Acacia tree that is situated in a crowded densely vegetated area, with several big trees standing close by. This presents challenge to fully or adequately take sample images of the tree. Tree_2 is tall, with wide-spread canopy, dense foliage and complex branch structure. Figure 1 shows the morphology of these two trees.

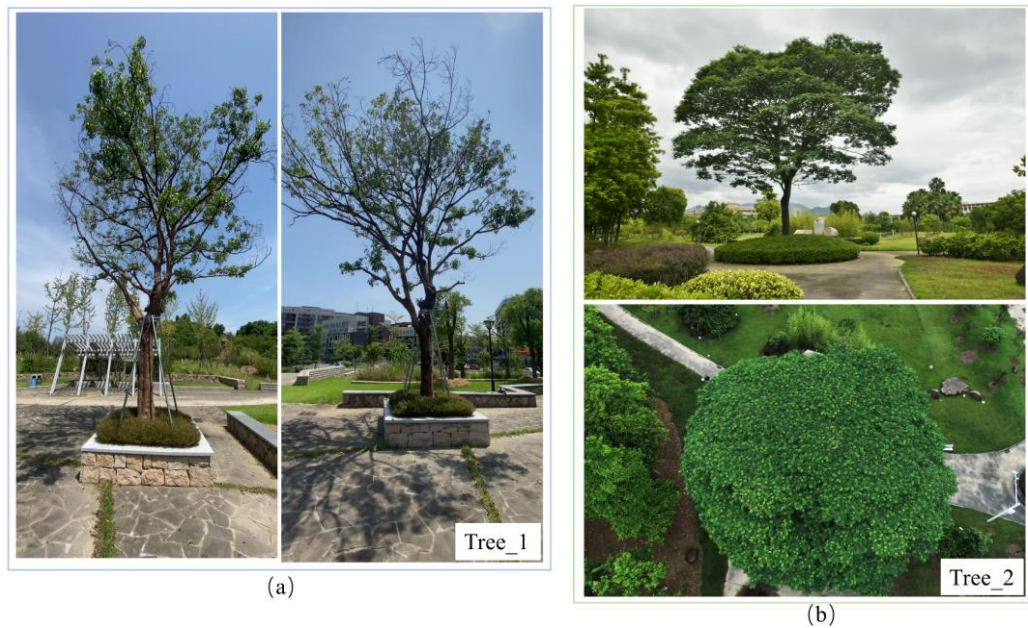

**Figure 1.** The structures and shapes of two trees used in this study. (a) is Tree_1 viewing from two perspectives on the ground and (b) is Tree_2 with views from both on the ground and in the air.

## 2.2. Research Methods

### 2.2.1. Traditional photogrammetric reconstruction

The development of photogrammetric reconstruction based on multiple overlapping images has been relatively mature by now, in which Structure from Motion (SfM) [7] and multi-view stereo (MVS) [18] are the two major processing steps. SfM originates from computer vision, it is the process of reconstructing 3D structure of the scene from its projections into a series of images taken from different viewpoints (motions of the camera). The result of SfM are the exterior and interior parameters of the camera, and sparse feature points of an object or scene. MVS, on the other hand, is designed to reconstruct a complete, dense 3D object model from a collection of images taken from known camera viewpoints.

This typical reconstruction process is implemented in many open source and commercial software including but not limited to COLMAP (https://colmap.github.io/), Agisoft Metashape (https://www.agisoft.com), and Pix4Dmapper (https://www.pix4d.com/pix4dmapper), among many others. We used the open-source COLMAP (Version 3.6) as the photogrammetric tool for reconstruction comparison experiments; and COLMAP is also used in NeRF to generate calibrated images for neural network training.

### 2.2.2. Neural Radiance Fields (NeRF) reconstruction

Neural Radiance Fields (NeRF) is able to implicitly represent a static object or scene with a multilayer perceptron (MLP) neural network, which is optimized to generate a picture of the scene from any perspective [11]. NeRF's superb ability to represent implicit 3D information has led to its rapid application in areas such as new perspective synthesis and 3D reconstruction. The principle is to represent a continuous scene implicitly by a function whose input is a 5D vector (3D position coordinates X= (x, y, z) and 2D view direction d= ($\theta$, $\phi$)), and the output is the color information c = (r, g, b) and the bulk density $\sigma$ about the point at X. In NeRF, the function is approximated with a MLP continuously optimized for the implementation. This function can be denoted as:

$$F_\theta : (X, d) \rightarrow (c, \sigma) \qquad (1)$$

The density and color information obtained from the MLP are then rendered using the volume rendering method to obtain the color information C(r), and finally the parameters of the MLP network are optimized by the Loss function between the voxel-rendered synthetic image and the real image to obtain the final implicit scene representation. Positional coding and hierarchical volume sampling methods are used to optimize the NeRF.

Although the initial NeRF method is concise and effective, it also suffers from problems such as long training time and aliasing artifacts. In order to solve these problems and improve the performance of the NeRF method, many new methods have been developed: the Instant-ngp [19] and Mip-NeRF [20] are two such notable examples. The most significant recent advancement is nerfstudio [21], which is a modular PyTorch framework for NeRF development that integrates various NeRF methods. Nerfstudio allows users to train their own NeRFs with some basic coding ability and knowledge and suitable hardware. It provides a clean web interface (viewer) to display the training and rendering process in real-time and can export the rendering results to video, point cloud and mesh data. Nerfstudio's default and recommended method is Nerfacto [22], which draws on the strengths of several methods to improve its performance. It uses a piecewise sampler to produce the initial set of samples of the scene, which makes it possible to sample even distant objects. These samples are then input to the Proposal Sampler proposed in MipNeRF-360 [23], which consolidates the sample locations to the regions of the scene that contribute most to the final render (typically the first surface intersection). In addition, Nerfacto combines a hash encoding with a small fused MLP (from Instant-ngp) to realize the density function of the scene, which ensures accuracy and high computational efficiency.

In summary, we used the nerfstudio framework for NeRF reconstruction experiment.

## 2.3. Data Acquisition and Processing

### 2.3.1. Data acquisition

Several tools were used for data collection. A smartphone camera was used to take photo and record video around the trees on the ground. In addition, for Tree_2 a Nikon digital camera was also used to collect photos on the ground, and a DJI Phantom 4 UAV was used to acquire image data from both on the ground and in the air. The numbers of acquired images are shown in Table 1. Considering the need for quality ground truth data, a RIEGL VZ-400 terrestrial laser scanner was used to perform multi-station scanning of the two target trees to obtain point cloud data for reference. Tree_1 and Tree_2 was scanned from 3 and 4 stations, respectively, with a scanning angular resolution of 0.04 degrees. The data were fine-registered in RiSCanPro（http://www.riegl.com/products/software-packages/riscan-pro/） with an accuracy of about 5 mm.

Table 1. Image data information sheet

| Image Dataset | Number of images | Image resolution | Total pixel (Millions) |
|---|---|---|---|
| Tree_1_phone | 118 | 2160×3840 | 979 |
| Tree_2_phone | 237 | 1080×1920 | 491 |
| Tree_2_nikon | 107/66 | 8598×5597 | 5149 |
| Tree_2_uav | 374 | 5472×3648 | 7466 |

Note: The first part of the dataset type name represents the target tree, and the last part describes the image collection sensor, which can be smartphone camera (phone), digital camera Nikon (nikon) or drone (uav). For example, Tree_2_uav refers to the data collected using a drone on Tree_2；For Tree_1_phone and Tree_2_phone, 118 and 237 frames were extracted from the recorded videos; For Tree_2_nikon, there were 107 images in total, but only 66 of these images can be calibrated after the SfM procedure.

### 2.3.2 Data processing

Photogrammetry and NeRF share the same first step of processing, which is SfM, which recovers image position and orientation, as well as generates sparse feature points of the scene. From here on photogrammetry uses MVS or other algorithms for dense point generation, while for NeRF the images and their positions and orientations were used for training and validating MLP model. Object and scene points can be generated and exported from nerfstudio; we then compare three sets of tree point cloud for their visual appearances and information contents after bringing both point cloud derived from photogrammetry and NeRF methods to the common coordinate system of the TLS point cloud and changed their scales. Fig. 2 shows the workflow of this study. All the image processing (photogrammetry and NeRF) were conducted in the same configuration setting of a cloud server platform; the computing equipment is equipped with Windows10 operating system, 12-core CPU, 28GB of RAM and NVIDIA GeForce RTX 3090 (24GB VRAM) GPU.

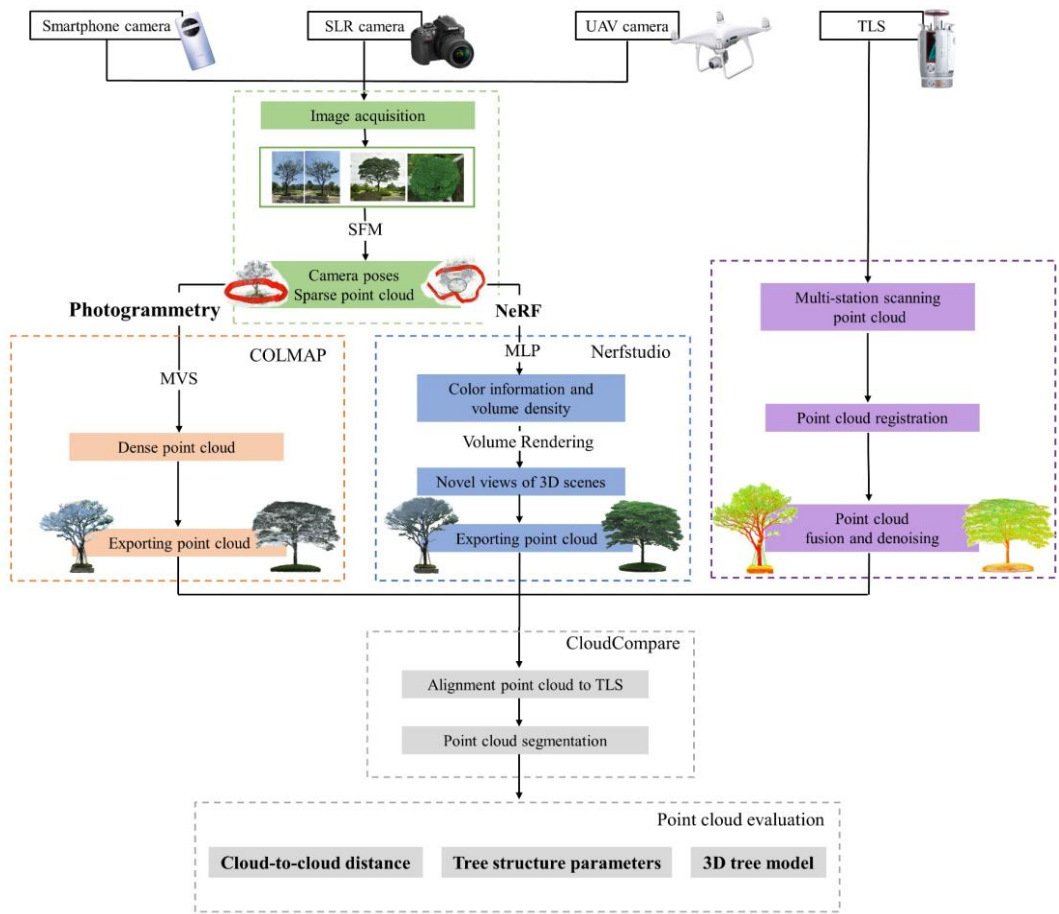

**Figure 2.** Complete workflow of the experiment

# 3. Results

## 3.1. Reconstruction Efficiency Comparison

Since SfM is a step utilized by both NeRF and photogrammetry methods, in a sense NeRF is similar to the MVS step in photogrammetry for generating dense points of the scene. So we compare the time used for COLMAP to carry out the densification process and the time NeRF used for training 10000-epoch neural network. We observed that usually after 7000 epochs of training the loss function usually converged, so 10000-epoch seems to be a conservative number. As shown in Table 2, COLMAP spent much more time than NeRF did to produce dense points of the scene. The time consumed by NeRF reconstruction always stays within the range of 10-15 minutes for different image sequences, while the time taken by COLMAP is 4 to 9 times longer and it increases with the number of images. For Tree_2_uav dataset, the COLMAP program failed to generate dense points after running for more than 127 minutes due to internal error.

**Table 2.** Computation time (minutes) of dense reconstruction for different trees and methods

|        | Tree_1_phone | Tree_2_phone | Tree_2_nikon | Tree_2_uav |
|--------|--------------|--------------|--------------|------------|
| COLMAP | 98.003       | 102.816      | 50.673       | failed     |
| NeRF   | 11.5         | 12.0         | 12.5         | 14.0       |

## 3.2. Point Cloud Direct Comparison

We imported the point clouds obtained from COLMAP and NeRF reconstruction into CloudCompare (https://www.cloudcompare.org/), finely aligned them with the TLS point cloud. We then extracted the targeted trees from the scene manually before conducting further comparative analyses. For each reconstructed tree point cloud model, we calculated its cloud-to-cloud distance to the corresponding TLS point cloud using CloudCompare.

**Table 3.** Number of points of the tree point cloud models

| Tree ID | Model ID            | Number of point |
|---------|---------------------|-----------------|
|         | Tree_1_Lidar        | 990,265         |
| Tree_1  | Tree_1_COLMAP       | 1,506,021       |
|         | Tree_1_NeRF         | 1,275,360       |
|         | Tree_2_Lidar        | 2,986,309       |
|         | Tree_2_phone_COLMAP | 1,746,868       |
| Tree_2  | Tree_2_phone_NeRF   | 1,075,874       |
|         | Tree_2_nikon_COLMAP | 580,714         |
|         | Tree_2_nikon_NeRF   | 1,986,197       |
|         | Tree_2_uav_NeRF     | 1,765,165       |

Note: The first part of the Model ID name represents the name of the target tree, and the middle part describes the image collection sensor, which can be smartphone camera (phone), digital camera Nikon (nikon) or drone (uav), and the last part denotes the image reconstruction method, which can be either COLMAP or NeRF. For Tree_1 the only image sensor is phone, so it is neglected in the naming.

As shown in Table 3, in most cases more than 1 million points can be reconstructed from images for each tree's 3D point cloud model, and COLMAP will produce more points than the corresponding NeRF. It also shows that COLMAP was not able to deal with the images taken by the Nikon for Tree_2 very well, indicated from the low number of points for Tree_2_nikon_COLMAP. And as mentioned in the previous section, COLMAP failed to dense reconstruct the Tree_2_uav.

Fig. 3 presents the three versions of point cloud models of Tree_1 and illustrates their differences. COLMAP and NeRF both reconstructed the target tree's trunk, canopy, and the surrounding ground scene with color completely, but both have white noise in their canopies. COLMAP's noise is distributed on the crown and trunk surfaces, while NeRF's is concentrated on the crown surface. To quantify the spatial distribution of the points, we calculated the cloud-to-cloud (c2c) distance after SOR denoising in CloudCompare. As shown in Fig. 3(e) and 3(f), in the color scale bar, we set all the distances greater than 0.4 m to be red. It can be seen that relative to the Lidar data, COLMAP has missing or distant tree parts at the top of the canopy, while NeRF has missing or distant tree elements in the middle of the canopy and at some branches. In addition, the mean cloud distance of the Lidar model with respect to COLMAP and NeRF models are 0.0305m and 0.0369m, with standard deviations of 0.0446m and 0.0574m, respectively.

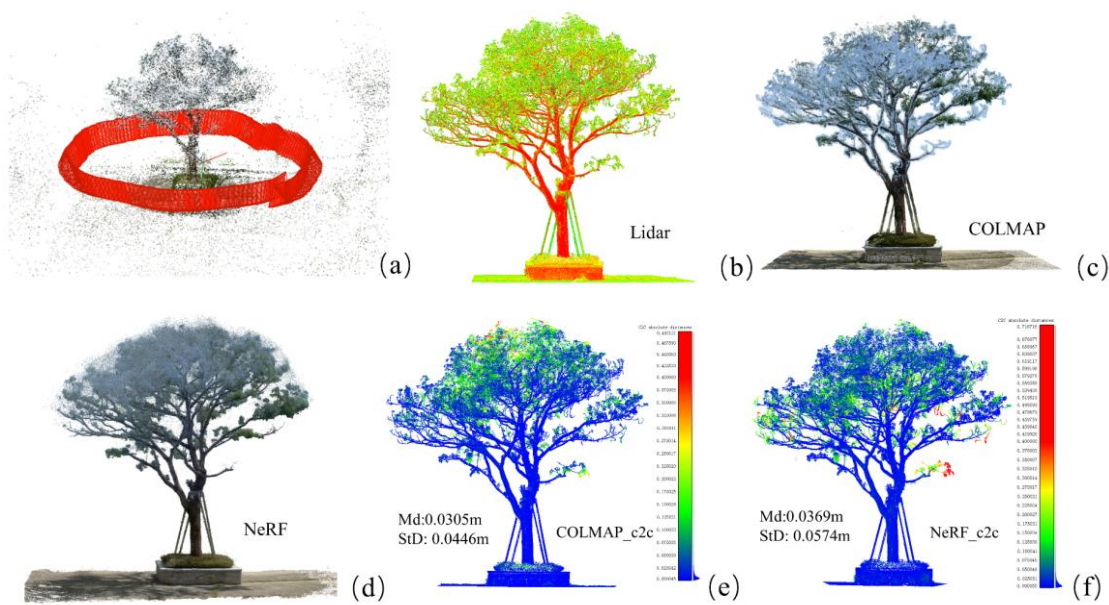

**Fig 3.** Tree_1 and its reconstruction result comparisons: (a) camera poses shown in red and sparse

points of the scene; (b) TLS Lidar point cloud , with intensity values colored in red and green representing trunks (branches) and leaves; (c) COLMAP model with color in RGB; (d) NeRF model, also color in RGB;(e) the cloud-to-cloud (c2c) distance between TLS Lidar and COLMAP model; (f) the c2c distance between TLS and NeRF model. Md is mean distance, StD is standard deviation.

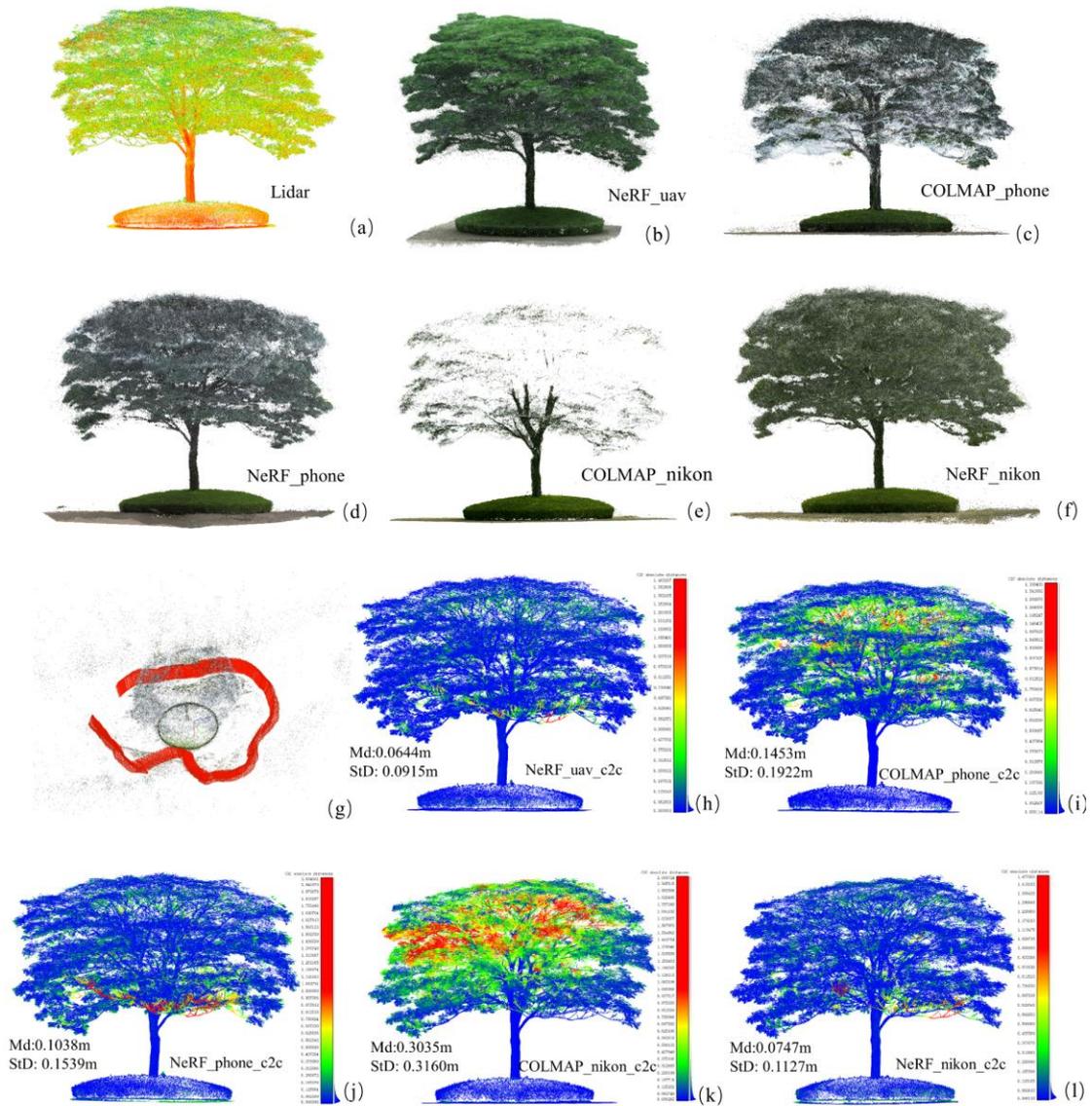

**Fig 4.** Tree_2 and its reconstruction result comparisons: (a) Lidar point cloud, with intensity values colored in red and green representing trunks (branches) and leaves; (b) NeRF_uav model with color in RGB; (c) COLMAP_phone model with color in RGB; (d) NeRF_phone model with color in RGB; (e) COLMAP_nikon model with color in RGB; (f) NeRF_nikon model, also color in RGB; (g) camera poses of the smartphone shown in red and sparse points of the scene; (h) the cloud-to-cloud (c2c) distance between TLS and NeRF_uav model; (i) the c2c distance between TLS and COLMAP_phone model; (j) the c2c distance between TLS and NeRF_phone model; (k) the c2c distance between TLS and COLMAP_nikon model; (l) the c2c distance between TLS and NeRF_nikon model. Md is mean distance, StD is standard deviation.

As shown in Fig 4 and Table 3, among the six versions of Tree_2 point cloud model, Lidar model has the largest number of points and the best quality. It is followed by the NeRF_uav model, which better reproduces the true color of the tree and has very little canopy noise. The COLMAP_nikon model has the worst quality, which only reconstructs the trunk portion of the tree, with the outer canopy shell partially recreated and more complex canopy interior portion missing. For COLMAP_phone and NeRF_phone models both reconstructed successfully from image frames taken from a smart phone video, the reconstruction quality is better even though the number of points is 700,000 less for NeRF method, which can reconstruct more dendritic information within the canopy with less noise. Similarly, we take the point cloud models of COLMAP and NeRF as a reference and calculated the cloud-to-cloud distance between them and TLS model. Then in the scale bar, we set the areas with distances greater than 1 m to be red, and assume that these parts are most likely to be missing. It can be seen that the NeRF_uav and NeRF_nikon models are the best in quality when compared to the LiDAR data, with only some missing branches in the lower part of the canopy, while the worst is the COLMAP_nikon model, with almost all missing branches and leaves inside the canopy. In addition, the mean cloud distances of the LiDAR model with NeRF_uav, NeRF_nikon and COLMAP_nikon are 0.0644m, 0.0747m and 0.3035m, with standard deviations of 0.0915m, 0.1127m and 0.3160m, respectively. The NeRF_phone model has more branches inside the canopy compared to the COLMAP_phone model, although some branches are missing in the lower part of the canopy, and the COLMAP_phone model has more missing elements inside the canopy. The mean cloud distances for NeRF_phone and COLMAP_phone are 0.1038m and 0.1453m, with standard deviations of 0.1539m and 0.1922m, respectively.

## 3.3. Extraction of structural parameters from tree point cloud

We extract the tree structural parameters from the point cloud derived from TLS and reconstructed from images via photogrammetry and NeRF using commercial and open source software Lidar360 (https://www.lidar360.com/)and 3DForest (https://www.3dforest.eu/), respectively. The parameters derived from laser scanning point cloud model are used as reference to compare with those derived from different reconstructed models.

The structural parameters of tree height (TH), diameter at breast height (DBH), crown diameter (CD), crown area (CA) and crown volume (CV) of a single tree were obtained after a series of processing steps such as denoising and single-tree segmentation in Lidar360 software, as summarized in Table 4. Due to limited or incomplete reconstructed canopy, Tree_2_nikon_COLMAP was not involved in the following processing.

Table 4. Results of Lidar360 extracted structural parameters for the studied trees

| Models | TH(m) | DBH(m) | CD(m) | CA($m^2$) | CV($m^3$) |
|---|---|---|---|---|---|
| Tree1_Lidar | 8.2 | 0.345 | 7.1 | 39.5 | 137.0 |
| Tree1_NeRF | 8.4 | 0.318 | 7.1 | 39.2 | 139.7 |
| Tree1_COLMAP | 8.1 | 0.349 | 7.0 | 38.9 | 138.0 |
| Tree2_Lidar | 13.6 | 0.546 | 16.3 | 208.8 | 1549.2 |
| Tree2_uav_NeRF | 14.0 | 0.479 | 16.0 | 201.0 | 1531.9 |
| Tree2_nikon_NeRF | 14.7 | 0.461 | 16.5 | 214.6 | 1627.0 |
| Tree2_phone_NeRF | 14.3 | 0.469 | 17.0 | 225.7 | 1693.1 |
| Tree2_phone_COLMAP | 13.8 | 0.562 | 16.7 | 220.3 | 1657.4 |

From Table 4, it can be seen that the structural parameters of both COLMAP and NeRF models for Tree_1 are not much different from those derived from the lidar model, used as ground truth here, with errors staying in a narrow range. COLMAP extracted parameters are closer to the lidar derived values in tree height, DBH and crown volume metrics, but NeRF extracted values have smaller error in crown diameter and crown area, which is also in line with the results of the visual comparisons. For Tree_2, both COLMAP and NeRF models have higher tree heights estimates than that of the lidar model, with phone_COLMAP being the one with the least error. In terms of DBH, again in agreement with the results of the visual comparisons, the DBH of the NeRF tree model was smaller than that of the Lidar model by a range of between 6.7cm and 8.5cm, whereas the DBH of COLMAP model was coarser than the Lidar model by 1.6 cm. And the errors of crown diameter are uniformly in the range of 0.2 m to 0.7 m. The uav_NeRF is closer to Lidar in terms of crown area and volume, and the model reconstructed using smartphone images deviates most from Lidar's values.

To compare different algorithms and test the usability of the derived point cloud, we also used 3DForest to extract certain important structural parameters, namely the tree height, DBH, and tree length (TL). The results are shown in Table 5:

**Table 5.** Extraction of major structural parameters of the models by 3DForest

| Models | TH(m) | DBH(m) | TL(m) |
|---|---|---|---|
| Tree_1_Lidar | 8.16 | 0.347 | 8.07 |
| Tree_1_NeRF | 8.32 | 0.321 | 7.95 |
| Tree_1_COLMAP | 8.02 | 0.346 | 8.04 |
| Tree_2_Lidar | 13.35 | 0.548 | 17.20 |
| Tree_2_uav_NeRF | 13.75 | 0.476 | 17.19 |
| Tree_2_nikon_NeRF | 13.55 | 0.468 | 18.0 |
| Tree_2_phone_NeRF | 13.89 | 0.502 | 18.22 |
| Tree_2_phone_COLMAP | 13.86 | 0.557 | 17.64 |

In Table 5, among the single tree model parameters extracted by 3DForest, DBH and TH remained consistent with those extracted by Lidar360, with the difference within the range of centimeter or even millimeter scale, which proved the reliability and accuracy of the extracted model parameters, as well as the quality of input data and robustness of the relevant processing algorithms.

From these structural metrics extracted from both Lidar360 and 3Dforest, it appears that from the NeRF point cloud model of trees the tree height tends to be over-estimated while the DBH under-estimated. This feature may be related to how the meshes and points were derived from the underlying volumetric scene representation.

## 3.4 Comparison of 3D tree models generated from point cloud

Based on the Lidar, COLMAP and NeRF reconstruction point cloud models, tree modelling was performed using the AdTree [24] open source project to generate 3D models of the target trees. Based on these 3D models, further comparative analyses can be performed to explore the usability and accuracy of point clouds generated by different reconstruction methods.

Similar to TreeQSM [25], AdTree extracts the tree skeleton and reconstructs a 3D branch model from the inputting tree point cloud. Figure 5 shows point cloud models, generated 3D branch models and the superimposed point cloud on branch models for detailed comparison. At first glance, all three branch models bear a high degree of resemblance to their corresponding point cloud models; but closer examination reveal more or less disconformities for each of them. Lidar point cloud has the best quality, but there are some branches that were not got reconstructed (marked in the blue box); NeRF's branch model looks messy in the tree crown with a lot of smaller branches; and COLMAP's crown looks better than that of NeRF's, but in COLMAP's branch model there is some distortion at the trunk, while NeRF's trunk is smoother (marked in yellow box). Merged point and branch model reveals that the trunks of all three branch models are a bit thicker than the point cloud models (marked in red boxes).

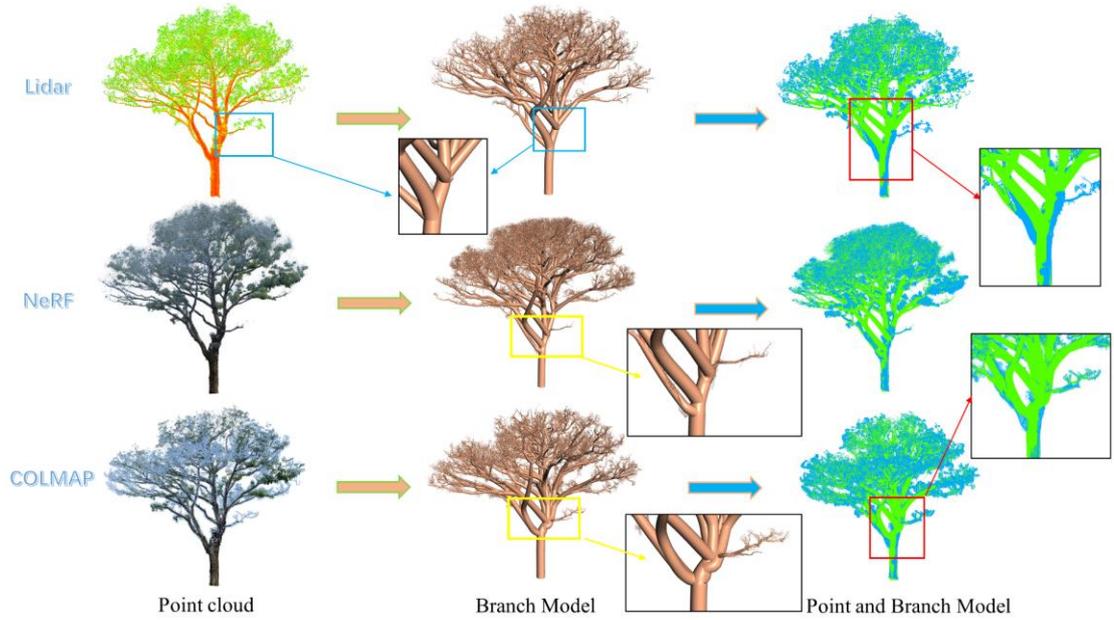

**Fig 5.** Tree_1 and its 3D branch models generated by AdTree. Left column is Point cloud model, with the LiDAR model colored with intensity values, the NeRF and COLMAP models colored in RGB; Middle column is Branch Model; Right column is merged point and branch model, with the Point cloud model color in blue and the Branch model in green; The color boxes are used to highlight the details of the model.

The AdTree generated 3D branch models for Tree_2 are shown in Figure 6. Overall tree height and tree morphology of the branch models are consistent with the point cloud models, except for the COLMAP_phone, which has substantial differences with the point cloud model in the canopy part. These upright-growing branches in the crown (marked in green box) are not found in the point cloud model. In addition, the trunk position and shape of the Lidar, NeRF_uav and NeRF_nikon models are similar, while those of NeRF_phone and COLMAP_phone are different (marked in blue boxes). NeRF_phone's trunk is misplaced from its position in the point cloud, while the latter's trunk is simply incorrect. Merged point and branch models demonstrate that the trunk and branch portion of the branch model is thicker than the point cloud model, where Lidar's and NeRF_uav's models are closer to the point cloud model, while the COLMAP_phone model has the largest deviation from the point cloud model (marked in yellow box).

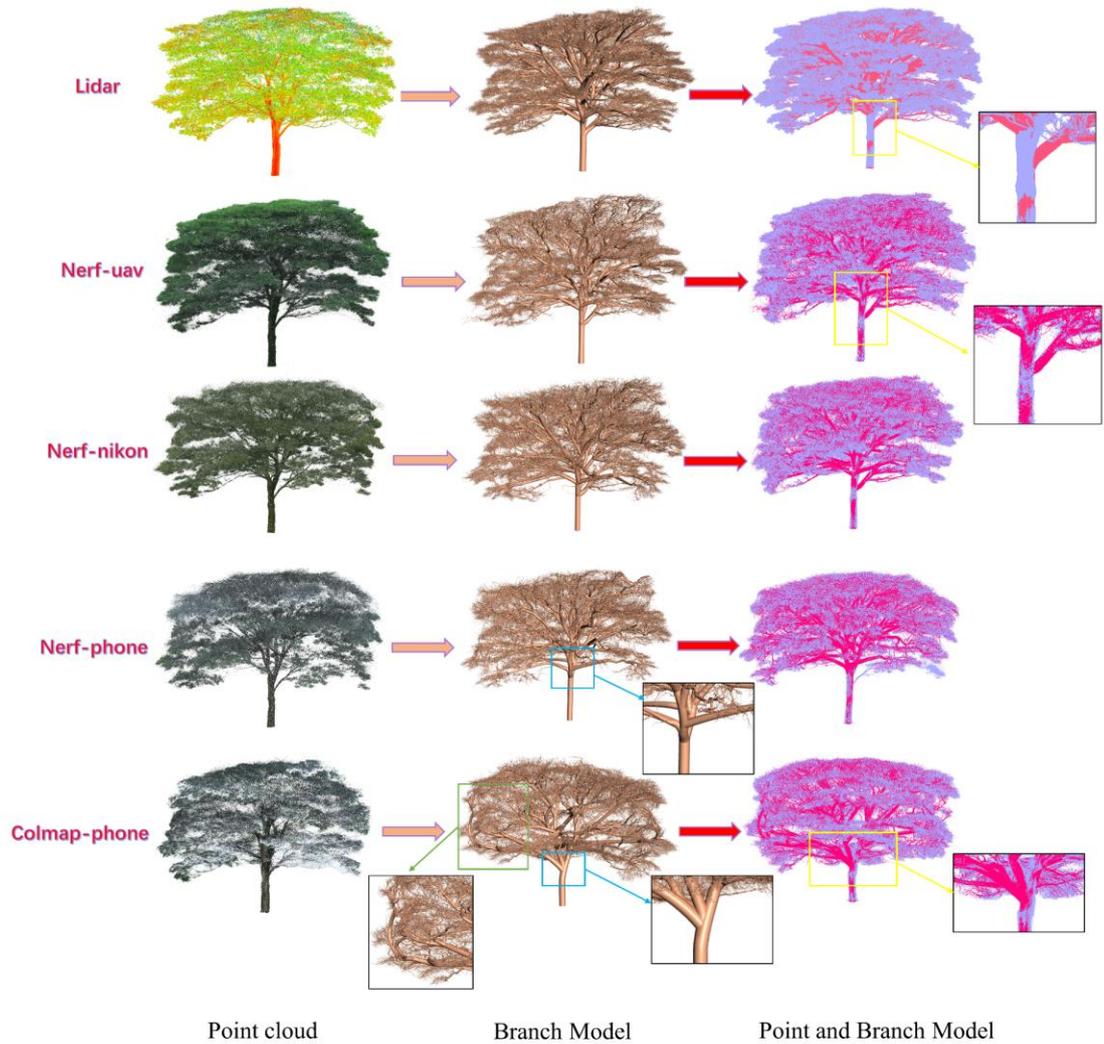

**Fig 6.** Tree_2 and its 3D branch model result reconstructed by AdTree. Left column is Point cloud model, the LiDAR model is colored with intensity values, the NeRF and COLMAP models are colored in RGB; Middle column is Branch Model; Right column is the merged point and branch model, with the Point cloud model color in purple and the Branch model in red; the color boxes show the portions enlarged to highlight the details of the model.

## 4. Discussion

In this study we compare two image-based 3D reconstruction methods for trees. One is based on the photogrammetry pipeline (SfM and MVS), while the other is based on the neural radiance fields (using a set of images with known camera poses, usually obtained from SfM procedure, to train and optimize a multi-layer perceptron network). Dense point cloud generated from these two methods was compared to the reference point cloud obtained from multi-station terrestrial laser scanning for reconstruction completeness and quality analysis.

The first observation is 3D reconstruction will not always be successful. There are certain minimum number of images requirement that must be met, but even for a large number of input images of trees and vegetative areas, the reconstruction can still fail. For Tree_2 in our study, in the few cases when images acquired from Nikon (107 images taken from the ground perspective) and drone (374 images taken both from the ground and in the air) were processed in COLMAP, partial reconstruction is achieved for Nikon images (only 66 photos calibrated after SfM), and no dense reconstruction for drone images. Concerned that open-sourced COLMAP might not be powerful enough, we also tested these two image datasets in the popular commercial photogrammetry software Metashape (Version 2.0.3) and Pix4dMapper (Version 4.7.5), getting the same failed reconstruction results. In contrast, NeRF method was able to handle these situations well even in the face of limited input images and still managed to produce visually appealing outcomes using less time.

MVS utilizes the output of SfM to compute depth or normal information for every pixel in the image, then fuses the depth and normal maps of multiple images in 3D to produce dense point cloud of the scene [27]. It is well recognized in computer vision that MVS face challenges when dealing with thin, amorphous, self-similar or shiny objects, and trees with dense leaves or thin branches are these types of objects and they are especially difficult to reconstruct using traditional photogrammetry method. NeRF can certainly play a role in filling the gap in this regard. Furthermore, MVS generally is a time-consuming process, sometimes it takes days to process dataset of a few thousand images. With NeRF it has the ability to greatly reduce the time required for reconstruction.

To ensure that the point cloud generated not only looking good, but also represent the object or scene with high fidelity, we evaluate the metrics of cloud-to-cloud distances. For Tree_1, those two c2c distances (Mean±STd) are very close: 0.031±0.045m for COLMAP model vs 0.037±0.057m for NeRF model; but for Tree_2, statistically NeRF generated point cloud models all have smaller cloud-to-cloud distances, meaning they have closer spatial distribution to the reference TLS point cloud than those of the point cloud generated from MVS; this higher spatial proximity proves that NeRF model can better represent the real scene in 3D.

Tree structural parameters extraction comparison illuminates the strength and weakness of these two image-based tree reconstruction methods. We focus our discussion on the five parameters extracted in Table 4. NeRF generated point cloud tend to be noisy, with large footprint or lower resolution. In terms of tree height and DBH, current photogrammetry method has advantages as they can offer results of higher accuracy; for NeRF models of two trees, the best parameters extracted have a relative error of 2.4-2.9% in tree height, 7.8 to 12.3% error in DBH. However, in terms of canopy properties such as canopy width, canopy area and canopy volume, NeRF method can provide better estimates than those derived from photogrammetry reconstruction models. And both types of point cloud can be fed into 3D tree modeling programs such as AdTree or TreeQSM to create branch models and further can be used to estimate tree volume and carbon stock.

We used various cameras to take images or shoot videos. Video seems to have an edge in guarantying successful reconstruction results, as both of our studied trees were able to be reconstructed by photogrammetric method using video frames collected by smartphone camera. It is more efficient to record videos than taking still photos, but the frames extracted from the video have lower resolution than normal images. For Tree_2, highly overlapped consecutive frames can be used to photogrammetrically reconstruct the tree, while images taken by other cameras (Nikon and UAV) even in similar manner (by shooting photos while walking or flying around the tree) but with less amount of overlap failed to accomplish that goal in COLMAP. NeRF doesn't have this restrictive overlap requirement, and the best quality NeRF-generated point cloud is trained from UAV images, followed by Nikon images, the worst being phone image frames. This demonstrates the importance of image resolution for NeRF reconstruction. Future field image collection may consider acquiring video of at least 4K resolution. When the number of images to be processed is large, NeRF-based methods will have the benefit of better efficiency. In addition, the weather and lighting condition can also affect the reconstruction and rendering result.

Compared with traditional reconstruction methods, NeRF is faster and has better reconstruction quality, especially for trees with dense canopies.

The result of NeRF rendering depend on the number of training epochs of the network, and it becomes stable after a certain threshold is reached. We investigated how many training iterations are needed in order for the network optimizer to converge, in part due to the fact that we need to pay hourly charge for using the cloud-based GPU. We determined that when the number of training epochs reaches 10000, the reconstruction results tend to stabilize. We may change the number of epoch, and tweak other system parameters to further examine factors that might affect reconstruction outcomes.

Similar to photogrammetry reconstructed model, models generated from NeRF methods are scale-less. We need to use markers, or ground control points to manually bring them to the real physical world, to make the reconstructed point cloud measurable. We noticed that the root mean squared error of the point cloud coarse and fine alignment is around 5cm in CloudCompare, and we are pondering ways to reduce this alignment error and improve point cloud accuracy.

Through this study we have better idea about the properties of NeRF-generated point cloud, and found answers to the questions raised in Introduction; in the meantime, we envision many further research plans in this area: how will the accuracy of the input parameters (3 positions and 2 directions) affect the precision of the result; new approaches to reduce noise, improve the detail and resolution of the underlying scene representation; how to distill the NeRF into geometrically-accurate meshes and point cloud; combination or fusion of NeRF, photogrammetry and laser scanning data; large-scale forest scene applications for plot- and landscape-level forest inventory information collection and analysis.

In this study, we found that NeRF 3D reconstruction of single tree can achieve remarkable performance, and we believe there is great potential for future application of NeRF to complex forest scenes. The technology is developing very fast, and we expect more sophisticated and powerful tools like Gaussian Splatting [28] will be coming soon to make 3D forest scene reconstruction more affordable and accessible.

## 5. Conclusion

In this research, NeRF technique is applied to 3D reconstruction of two trees with different canopy structure and compared with the traditional photogrammetric reconstruction technique. We tested the processing efficiency and capabilities of these two competing approaches using a series of images of trees acquired with different cameras and viewing angles. Quantitative and qualitative analyses were conducted to examine the visual appearance, reconstruction completeness, information content and utility of the NeRF- and photogrammetry- generated point cloud. Specifically, we looked into metrics including cloud-to-cloud distance, tree structural parameters and 3D models for comparison. The results show that:

(1) The processing efficiency of the NeRF method is much higher than that of the traditional photogrammetric densification method of MVS, and it also requires less images for reconstruction.

(2) For trees with sparse or little leaves, both methods can reconstruct accurate 3D tree models; for trees with dense foliage, the reconstruction quality of NeRF is better, especially in the tree crown area. NeRF models tend to be noisy though.

(3) The accuracy of the traditional photogrammetric method is still higher than that of the NeRF method in the single tree structural parameters extraction in terms of tree height and DBH; NeRF models are likely to overestimate the tree height while underestimate DBH. However, canopy metrics (canopy width, height, area and volume and so on) derived from NeRF model are more accurate than those derived from photogrammetric model.

(4) The method of image data acquisition, the quality of images (image resolution, quantity), and the photographing environment all have an impact on the accuracy and completeness of NeRF and photogrammetry reconstruction results.

Further research is needed to determine the best practices.

Author Contributions: Conceptualization, H.H. and T.G.; methodology, H.H.; software, T.G.; validation, H.H. and T.G.; formal analysis, H.H. and T.G.; investigation, H.H. and T.G..; resources, H.H. and C.C.; data curation, H.H. and T.G.; writing—original draft preparation, H.H. and T.G.; writing—review and editing, H.H. and T.G.; visualization, T.G.; supervision, H.H.; project administration, H.H.; funding acquisition, C.C. All authors have read and agreed to the published version of the manuscript.

Funding: This research received no external funding.

Data Availability Statement: Data will be available upon reasonable request.

Acknowledgments: We would like to thank Mr. Cheng Li, Ms. Luyao Yang for their assistance in collecting and processing terrestrial laser scanning data of Tree_2 in campus.

Conflicts of Interest: The authors declare no conflict of interest.